\documentclass[10pt, a4paper]{article}

\usepackage[final]{lrec2026}

\usepackage{listings} 
\usepackage{makecell}
\usepackage{longtable}
\usepackage{booktabs}
\usepackage{float}
\usepackage{tabularx}

\title{\textbf{CaresAI at CT-DEB'26: Detecting Dosing Errors In Clinical Trials Using Domain-Specific Transformer Embeddings and Classification Models}}

\name{
\begin{tabular}[t]{c}
Leon Hamnett$^{1}$,
Favour Igwezeke$^{1,2}$,\\
Joseph Itopa Abubakar$^{1}$,
Mary Adetutu Adewunmi$^{1,3}$
\end{tabular}
}

\address{
$^{1}$CaresAI, Australia; 
$^{2}$Faculty of Pharmaceutical Sciences, Nsukka Enugu, Nigeria;\\
$^{3}$Menzies School of Health Research, Australia\\
\texttt{Leon.hamnett@caresai.org, Favour.igwezeke.249461@unn.edu.ng}\\
\texttt{Joseph.itopaa@gmail.com, Mary.adewunmi@menzies.edu.au}
}

\abstract{
Medication errors, particularly dosing errors in clinical trials (CT), can lead to patient harm, adverse drug events and worse patient outcomes. Dosing errors are preventable, and early identification can improve trial integrity and mitigate subsequent clinical and financial burden. This study aims to detect dosing errors within CT protocols by evaluating text representations of trial information using transformer-based language models trained on biomedical corpora. CT textual data was encoded using several models, including ClinicalBERT, PubMedBERT, BioBERT, and MedCPT, and integrated with categorical features. These text embeddings were used as input to classical machine learning models and neural network architectures within an experimental framework. Performance was primarily assessed using ROC-AUC with respect to predicting dosage error. Under a logistic regression baseline, BioBERT consistently outperformed alternative encoders, achieving an ROC-AUC of 0.794, a 3.95\% improvement over the ClinicalBERT baseline. Combining multiple embeddings did not yield improvements, indicating that domain alignment outweighs representational stacking. Gradient boosting models, support vector classifiers, logistic regression, and residual neural networks achieved the strongest performance for predicting dosage error, achieving ROC-AUCs: 0.821 to 0.853. Overall, the integration of domain-specific transformer embeddings with structured metadata enables discrimination of trials meeting a predefined elevated dosing error risk criterion, advancing safety monitoring and supporting informed regulatory decision-making.
 \\ \newline \Keywords{Medication Errors, Clinical Trials,Dosing Errors, Transformer Models,Natural Language Processing (NLP), Machine Learning}
 }

\begin{document}

\maketitleabstract

\pagestyle{empty}

\thispagestyle{empty}

\section{Introduction}
Clinical trials provide the structured and rigorously controlled framework required to evaluate the safety and efficacy of novel pharmaceutical compounds in human populations. However, this process is both costly and uncertain. Only 14\% of compounds entering clinical trials ultimately receive regulatory approval \citep{wong2019estimation}, despite substantial financial investment at each phase. Median costs are estimated at USD 3.4 million for Phase I, USD 8.6 million for Phase II, and USD 21.4 million for Phase III studies \citep{martin2017clinicalcost}. 

Beyond financial risk, trial protocols are inherently complex, involving numerous interacting variables, multiple treatment arms, and strict procedural constraints. This complexity increases the likelihood of operational and methodological errors, which may compromise statistical validity, delay development timelines, or invalidate findings entirely, resulting in wasted resources and lost therapeutic opportunity \citep{buyse2017impact,berger2009framework}.

Medication errors are unintended failures in the drug treatment process that cause, or have the potential to cause, patient harm \citep{ema_medication_errors}. These errors can include: incorrect timing of doses, incorrect dosage, additional unintended doses, or inappropriate administration rates \citep{tariq2024medication}, as well as errors in randomisation, masking, and broader experimental design \citep{berger2009framework}. Medication errors remain one of the most frequently occurring yet preventable causes of adverse patient outcomes within clinical settings \citep{tariq2024medication}. The same medication errors that occur in clinical settings can also arise during clinical trials. Medication errors during clinical trials can lead to reduced efficacy of the trial, serious toxicity, compromised patient safety, adverse drug events, and can also pose a threat to internal validity and interpretability of clinical trial outcomes \citep{ema_medication_errors}. 

Given the substantial consequences of erroneous approval decisions related to new compounds, rigorous monitoring and statistically robust evaluation are essential during the entire clinical trial process. Regulatory bodies such as the Food and Drug Administration (FDA) and European Medicines Agency (EMA), emphasise the importance of proactively identifying, monitoring and mitigating dosing errors during clinical trials due to their potential to contribute to a public health burden \citep{stewart2021fda}. In addition, timely identification of dosing errors in clinical trials can reduce both financial costs and patient risk by facilitating earlier correction or termination of ineffective studies \citep{rim2024evidence}.

However, clinical trial documentation often consists of large volumes of unstructured textual data, making the extraction of relevant methodological details time-consuming and challenging \citep{chen2020nlpclinicaltrials}. This complicates verification of trial integrity, particularly with respect to correct dosing and administration schedules across study arms and cohorts. The predominance of unstructured textual documentation within clinical trial reports and protocols, highlights the need for automated methods capable of systematically extracting and verifying critical methodological details. 

In the context of the medical field and more specifically when evaluating clinical reports, Natural Language Processing (NLP) provides a scalable approach for identifying and structuring relevant information embedded within narrative clinical text. Natural language processing is a subfield of computer science and artificial intelligence, which allows computers and digital devices to recognise and understand text and speech data \citep{ibm_nlp_2026}. Beyond information extraction from text, NLP techniques also support downstream analytical tasks such as diagnosis explanation, disease progression modelling, and treatment effectiveness assessment. Recent advances in machine learning have led to the development of increasingly sophisticated NLP algorithms with broad applications in information extraction and summarisation \citep{khurana2023nlp}. These developments within the field of NLP are particularly relevant to medical documentation, where clinical notes and reports generated from medical practise and research typically take the form of large volumes of unstructured text. NLP methods have already been applied within the medical field for some use-cases, including: extracting dosage information and administration schedules \citep{lu2016nlpmedication, mcneer2019postprocessing}, identifying clinical and treatment events \citep{bitterman2023endtoend}, and detecting adverse drug events and reactions \citep{murphy2023adverse,naderiandrug}. Recent research has also investigated specific applications of these NLP methodologies within the context of clinical trials, including: automating report drafting, extracting drug-to-drug interactions (DDI) and optimising protocol design in CTs \citep{chopard2021textmining, kuo2025ragllm, lee2024eligibility,zhu2020ddi}.

The rapid proliferation of transformer-based language models has significantly advanced NLP research \citep{kalyan2021ammussurveytransformerbased,patwardhan2023transformers}. These models leverage self-supervised learning and attention mechanisms to learn feature-rich, general-purpose representations from large volumes of unlabelled text corpora. As a consequence of extensive pre-training on a large amount of data, only relatively small amounts of task-specific data are required to achieve strong performance on downstream tasks, since the pre-trained parameters within each model already encode substantial linguistic and semantic structure derived from the training corpora.

During pre-training, transformer architectures learn contextualised representations of input text, capturing semantic and syntactic relationships within fixed-length numerical embeddings. Owing to their success in general-domain NLP, these models have increasingly been applied in the medical domain across tasks such as question answering, named entity recognition (NER), text summarisation, classification, and sentiment analysis \citep{nerella2024transformers,cho2024taskspecific}.

More recently, research has investigated whether domain-specific pre-training on biomedical corpora, such as PubMed abstracts and Mayo Clinic electronic health records, can further improve performance in a medical context \citep{WANG201812}. Empirical findings suggest that models pre-trained predominantly on medical text corpora achieve superior results on clinical tasks \citep{WANG201812}. Exposure to specialised clinical terminology and domain-specific vocabulary enhances representational precision, while repeated exposure to clinical narratives allows models to implicitly capture the structural patterns of medical documentation. Moreover, domain-aligned pre-training reduces distributional shift between pre-training data and downstream data, improving robustness and sample efficiency. This is particularly important in clinical contexts, where annotated datasets are scarce, costly to produce, or limited in scope.

The strong performance of pre-trained transformer models in NER tasks enables reliable extraction of key clinical entities, including diseases, medications, procedures, and intervention attributes.  This capability is especially relevant for clinical trial analysis, where experimental design elements, such as study arms, intervention types, and dosing schedules, are frequently recorded as unstructured text. There has already been some success in applying pre-trained transformer models within the context of clinical trials, including tasks related to: extracting outcome and arm information \citep{WHITTON2023102661} , distinguishing adverse event reporting \citep{chopard2021textmining}, and extracting drug–drug interactions \citep{zhu2020ddi}. However, despite these recent advances, access to large, openly available clinical trials datasets remains limited, restricting large-scale experimentation and hypothesis testing \citep{navar2016openaccess}. Recent initiatives such as TrialBench have sought to address this gap through multi-modal dataset construction and standardised baselines for clinical trial prediction tasks \citep{chen2025trialbench}.

Consequently, transformer-based models provide a scalable framework for structuring and validating critical methodological information embedded within clinical trial documentation. As part of this research project, the authors investigated the suitability of various transformer-based models which have been pre-trained on biomedical text corpora, in order to encode the textual information related to each clinical trial, and allow machine learning models to make predictions on whether a specific clinical trial contains a dosage error within its approach.

\section{Methodology}

\subsection{Dataset and Problem Formulation}

As part of the CT-DEB'26 challenge (Detecting Dosing Errors from Clinical Trials) \citep{ferdowsi2026ctdeb}, a dataset was generated  \cite{Heche2026EarlyRiskStratificationDataset}, containing structured and unstructured information for a large number of different clinical trials (written in the English language), collected from the ClinicalTrials.gov site \citep{clinicaltrials_gov}. Each row corresponded to a single clinical trial study and included long-form textual fields as well as several numerical and categorical variables related to each specific trial. As part of the dataset construction process, a binary label was generated: a label of 1 was assigned if the lower bound of the 95\% Wilson confidence interval for the dosing error rate exceeded 0.01\% \citep{ferdowsi2026ctdeb,Heche2026EarlyRiskStratificationDataset}.

The objective of this research was to compare several pre-trained biomedical models for encoding text features, and to develop and evaluate machine learning models (both classical and neural networks) capable of predicting which clinical trials met this elevated dosing error risk criterion. The study systematically evaluated the impact of different feature subsets, alternative input representations of textual data, and varying model architectures on predictive performance.

Given the presence of substantial unstructured text, it was necessary to transform textual data into machine-readable representations. To this end, several pre-trained transformer-based models were evaluated, including ClinicalBERT \citep{huang2020clinicalbert}, BioBERT \citep{lee2020biobert}, PubMedBERT \citep{pubmedbert}, and MedCPT \citep{jin2023medcpt}. These models, pre-trained on large-scale biomedical corpora, can generate contextualised vector representations of input text that capture domain-specific semantic information. The resulting embeddings were used as input features for downstream machine learning classifiers, including both traditional machine learning models, as well as more advanced neural networks. A controlled experimental framework was applied throughout the research process, modifying a single variable at a time while holding others constant to isolate individual effects of experimental changes. Standard machine learning practice was followed by using a separate subset of data to analyse the predictive performance of each model, to prevent bias by showing the model data it has already seen before during the training process.

The dataset was provided in two separate subsets: a training dataset of 29,478 data points and a hold-out validation dataset of 6,316 data points. Each dataset contained the following columns:

\begin{flushleft}
\texttt{phases, enrollmentCount, allocation, interventionModel, primaryPurpose, masking, healthyVolunteers, sex, oversightHasDmc, briefSummary, detailedDescription, conditions, conditionsKeywords, protocolPdfText, numArms, armDescriptions, armGroupTypes, numInterventions, interventionTypes, interventionDescriptions, interventionNames, numLocations, locationDetails, target, nctid.}
\end{flushleft}

A detailed description of the dataset variables can be found in Appendix \ref{appendix: columns}.
The dataset contained a mixture of categorical, numerical, and free-form text variables, each related to a specific clinical trial. 

In the final modelling pipeline, textual features (briefSummary, detailedDescription, armDescriptions, and interventionDescriptions) were encoded using BioBERT embeddings. Structured categorical variables, including allocation method, intervention model, primary purpose, oversightHasDmc, and trial phase, were one-hot encoded and concatenated with the text embeddings. Continuous numerical variables such as enrollmentCount, numArms, and numLocations were not included in the final models. The numArms variable demonstrated a weak statistical association with the target variable, while enrollmentCount and numLocations were not considered likely to provide meaningful predictive signal beyond the structured and textual features already incorporated.

The protocolPdfText field was not incorporated into the modelling pipeline due to substantial data sparsity and extreme variability in length. In the training set, 81.3\% of records (23,963 of 29,478) contained no extracted PDF text, whereas only 3.2\% of validation records were empty. This imbalance raised concerns that the model could implicitly learn the presence of non-empty PDF text as a proxy for dataset partition, thereby introducing distributional bias. Furthermore, when data was available within the protocolPdfText field, the extracted protocol text was extremely long (training max: 11,298,927 characters; validation max: 6,424,120 characters), with highly skewed distributions and very large standard deviations. Encoding such long documents within a fixed transformer context window would require aggressive truncation, potentially discarding the majority of the content and introducing inconsistent representation quality across samples. Given the high proportion of missing entries in training data, the substantial computational burden of processing very long documents, and the risk of introducing noise through heavily truncated representations, this feature was excluded from the final experiments. 

As the primary objective of this study was to evaluate the comparative effectiveness of pretrained biomedical language models for encoding unstructured clinical trial documentation, the modelling framework was designed to isolate the contribution of textual representations. Accordingly, only categorical structured features were included alongside the embeddings, and continuous numerical variables were excluded. 

\subsection{Overview of Experimental Investigations}

An iterative and controlled experimental framework was adopted. At each stage, only one variable or modelling component was modified while all other elements were held constant. This enabled the isolation of the effect of individual design decisions on predictive performance.

The experimental process was conducted in two major phases:\\
Phase 1: Iteration of text encoding methods and feature selection\\
Phase 2: Iteration of model architectures and training processes

\subsubsection{Phase 1: Text Encoding and Importance of Features}

Categorical Feature Analysis: Initial experiments employed a logistic regression classifier as a baseline model. Using this fixed architecture, the study evaluated the incremental impact of incorporating one-hot encoded categorical variables. Associations between categorical features and the target variable were assessed using Chi-squared tests and Cramér’s V statistics. Importantly, these statistical analyses were conducted exclusively on the training data to prevent information leakage, ensuring that the validation set remained fully unseen during both feature selection and model evaluation to give unbiased reporting of the models performance on unseen data. 

An ablation study was conducted by incrementally adding individual features to measure their contribution to predictive performance. Categorical features were selected to be used in the following stages, primarily based on statistical significance in relation to the target feature.

Text Encoding with Pre-trained Transformer Models: Given the presence of substantial unstructured text, multiple pre-trained biomedical transformer encoders were evaluated to generate contextual embeddings for the following textual columns: "Brief summary", "Detailed description", "Arm descriptions", "Intervention descriptions". The following encoders were compared: ClinicalBERT, PubMedBERT, BioBERT, MedCPT, and also combinations of multiple encoders. A short description of each transformer model is given below:

\begin{itemize}

\item \textbf{BioBERT} is a domain-specific language model pre-trained on large-scale biomedical literature, including PubMed abstracts and PubMed Central full-text articles, enabling it to capture biomedical terminology and contextual relationships in clinical text \citep{lee2020biobert}.

\item \textbf{ClinicalBERT} is a BERT-based model pre-trained on clinical notes from intensive care unit records in the MIMIC-III database, making it well suited for clinical narrative understanding tasks \citep{huang2020clinicalbert}.

\item \textbf{MedCPT} is a contrastive learning-based biomedical language model trained on 255 million user click logs from PubMed, designed to produce high-quality semantic embeddings for zero-shot biomedical information retrieval \citep{jin2023medcpt}.

\item \textbf{PubMedBERT} is a language model pre-trained from scratch exclusively on PubMed abstracts, without general-domain pre-training, allowing it to develop vocabulary and representations specifically optimised for biomedical text \citep{pubmedbert}.
\end{itemize}

Each pre-trained encoder generated fixed-length vector representations for the textual fields. A maximum sequence length of 256 tokens was selected as a practical compromise between contextual coverage and computational efficiency. Token-level analysis using the BioBERT tokenizer (see Appendix \ref{appendix:token_length}) indicates that, under this constraint, the proportion of truncated entries was 8.65\% for briefSummary, 31.81\% for detailedDescription, 7.79\% for armDescriptions, and 4.03\% for interventionDescriptions.Notably, the detailedDescription field contained a substantial proportion of empty entries 39.4\% in the training set (11604 of 29478 rows) and 34.6\% in the validation set (2187 of 6316 rows), which reduces the effective truncation impact for that field. Overall, more than 90\% of entries in briefSummary, armDescriptions, and interventionDescriptions were fully preserved within the 256-token limit; detailedDescription experienced higher truncation (31.81\%), though a substantial proportion of entries were empty. A uniform maximum token length was applied across all textual fields to ensure consistency within the encoding pipeline. While increasing the context window may allow additional long-range information to be captured, particularly for the detailedDescription field, this would substantially increase computational cost and memory requirements. The selected limit therefore reflects a trade-off between representational completeness and practical feasibility. 

Sentence-level embeddings were obtained using mean pooling over the token-level representations produced by each model. These embeddings were then used as input features for a downstream logistic regression classifier, which learned to map the numerical vector representations to the binary dosing error label. Comparative experiments evaluated the impact of different pre-trained encoders on validation ROC-AUC performance. BioBERT achieved the strongest results and was therefore selected as the primary text encoder for subsequent experiments.

\subsubsection{Phase 2: Iteration of Model Architectures and Training Processes}

The final set of inputs features was selected based on statistical analysis and domain knowledge, by combining BioBERT embeddings for all text fields and one-hot encoded categorical features (specifically: intervention model, allocation type, oversight, primary purpose, phase). These features were features that gave the model the most discriminative predictive power in determining if the probability of a dosage error having occurred within a specific clinical trial exceeded the defined threshold.

As the dataset exhibited substantial class imbalance (approximately 95.4\% of training records labelled as no dosage error - 28,126 of 29,478 data-points), it was necessary to investigate methodologies for mitigating this large imbalance. A series of experiments evaluated the impact of varying class weights during training. By varying the class weights, we can instruct a machine learning model to place greater importance on examples from a certain class during the training process. These experiments consisted of the following stages: no weighting (all examples assigned equal importance), fixed weighting ratios (1:5, 1:10, 1:20) toward the minority class, and automatically computed class weights based on the proportional amount of each element within the dataset (around 0.5:11 weighting, for 0 and 1 dosage error variables, respectively). 

Once the final set of input features was fixed, and an optimal class weighting had been identified, several experiments were undertaken to vary the type of machine learning model that was used to make predictions, and thus change the manner in which the model is learning from the input data. These models included both classical machine learning models and more advanced neural network architectures. The classical machine learning models that were examined were as follows: Logistic Regression, Linear Support Vector Classifier, Support Vector Classifier, K-Nearest Neighbours Classifier, Naïve Bayes, Decision Trees, XGBoost and LightGBM. Due to the high dimensionality of transformer embeddings, some classical models experienced convergence or computational limitations. Comparative evaluation identified Logistic Regression, SVC variants, as well as Gradient Boosting models as the strongest classical baselines. The top performing models had hyperparameter optimisation performed using GridSearch to identify the combination of model parameters that gave the best predictive power.

Following the evaluation of the classical machine learning models, multiple neural network architectures were explored: simple multi-layer perceptron (MLP), wide multi-layer perceptron, residual neural network (with skip connections), text-tower architecture (separate processing streams for text and categorical features, merged at a later part of the neural network). Specific implementation details of the various neural network architectures can be found in Appendix \ref{appendix: neural networks}.

Performance differences between neural network architectures was analysed using consistent feature representations and with other neural network training parameters fixed (e.g. loss, optimiser, learning rate, training epochs). During the later stages of investigation into neural network models, once the best performing architecture was identified, the training process itself was further optimised.  This included the following methodologies: early stopping based on validation ROC-AUC, learning rate scheduling (ReduceLROnPlateau, OneCycleLR), loss function comparison (binary cross-entropy vs focal loss), and model check-pointing to retain best parameters against the validation data. These experiments aimed to optimise convergence behaviour during model training and reduce over-fitting to the training data. All models were evaluated on held-out validation data to assess generalisation performance on unseen data. The primary evaluation metric was ROC-AUC, with secondary metrics used to characterise performance under class imbalance conditions. 

Extensive hyperparameter optimisation of the neural network models was not performed. Instead, the architectures were constructed to represent common and practically relevant structural variations, such as wide networks, residual connections, and gated text towers, rather than to exhaustively tune each configuration for maximal performance. Although more comprehensive optimisation could potentially improve absolute predictive performance, this was beyond the scope of the present study. The reported results therefore reflect controlled and representative configurations designed to enable a fair and consistent comparison across modelling approaches.

\subsubsection{Phase 3: Final Modelling}
Following identification of the strongest-performing classical models and neural network architectures, a final evaluation stage was conducted to obtain a more robust estimate of predictive performance. The original training and validation datasets were combined, and stratified k-fold (k=5) cross-validation was applied to create unique data subsets while preserving the overall proportion of positive and negative cases within each fold. 

In each iteration, the model was trained on four folds and evaluated on the remaining fold, producing five independent sets of performance metrics. Averaging these results provides a more reliable estimate of generalisation performance by reducing dependence on any single train–validation partition. Stratification further ensures that class imbalance is consistently represented across folds, thereby limiting variability caused by uneven class distributions and yielding a more stable and unbiased assessment of model performance.

\section{Discussion of Results}

\subsection{Results for Encoding of Text Columns}

\begin{figure}[h]
    \centering
    \includegraphics[width=1.0\linewidth]{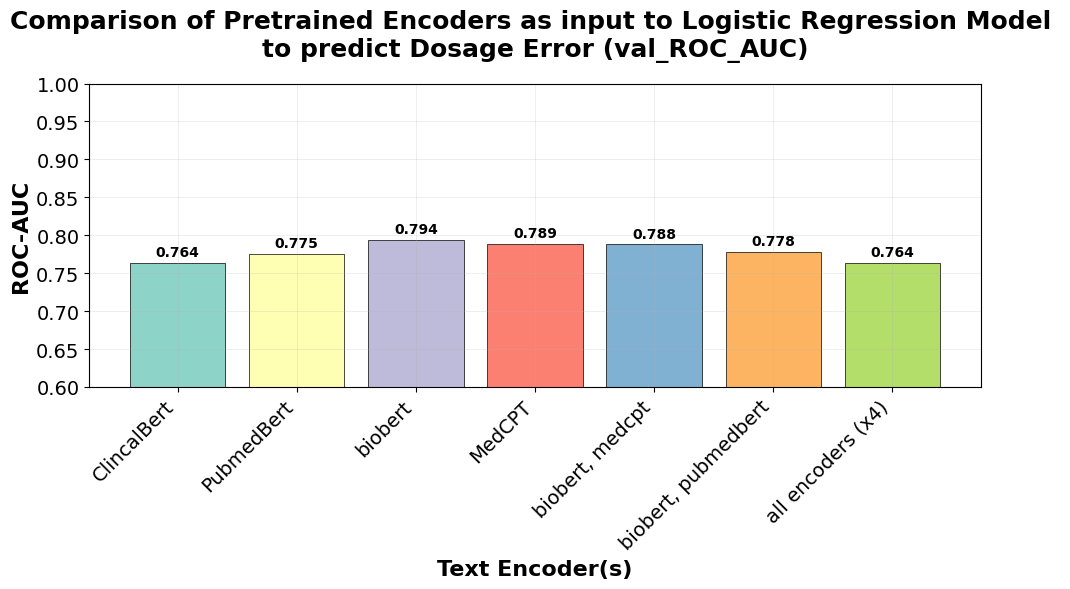}
    \caption{Comparison of pretrained encoder models (absolute \texttt{ROC-AUC})}
    \label{figure 1:text_encoding_comparison_abs}
\end{figure}

\begin{figure}[h]
    \centering
    \includegraphics[width=1.0\linewidth]{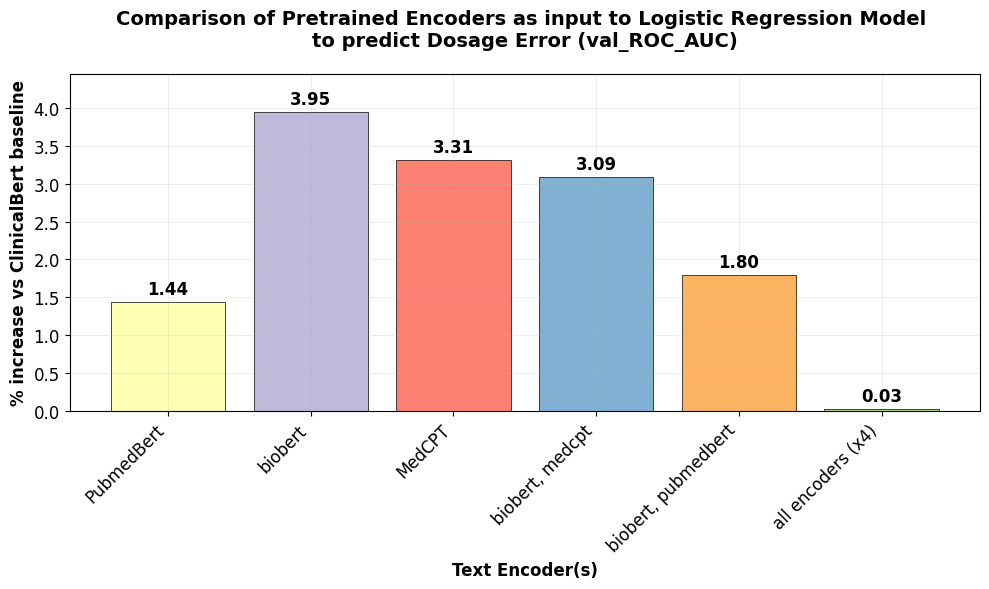}
    \caption{Comparison of pretrained encoder models (\% \texttt{ROC-AUC} improvement over ClinicalBert baseline)}
    \label{figure 2:text_encoding_comparison_percent}
\end{figure}

BioBERT demonstrated the strongest individual encoder performance, achieving a ROC-AUC of 0.794 against the validation set under logistic regression, compared to 0.775 for PubMedBERT, 0.789 for MedCPT, and 0.764 for ClinicalBERT (see Figure \ref{figure 1:text_encoding_comparison_abs}). This is consistent with the design of BioBERT, which was pre-trained on large-scale biomedical literature, including PubMed abstracts and PMC full-text articles, enabling it to capture domain-specific linguistic patterns relevant to clinical trial intervention descriptions. As shown in Figure \ref{figure 2:text_encoding_comparison_percent}, when compared to the ClinicalBERT baseline encoding method, BioBERT shows close to a 4\% improvement in ROC-AUC. ClinicalBERT, while pre-trained on clinical notes from intensive care unit records, may be less suited to the structured protocol language characteristic of ClinicalTrials.gov records \citep{alsentzer-etal-2019-publicly}. The more advanced model architectures, including XGBoost, LightGBM, and residual neural networks, achieved stronger predictive performance when using BioBERT embeddings, with ROC-AUC values ranging from 0.821 to 0.832. These results suggest an enhanced capacity to learn from the BioBERT vector representations and improve discriminative ability in identifying trials meeting the predefined elevated dosing error risk threshold.

Contrary to the initial expectation that combining multiple encoders would yield additive gains, results under logistic regression showed that the combined four-encoder representation (ROC-AUC: 0.764) under-performed BioBERT alone, showed minimal difference in ROC-AUC compared to the ClinicalBERT baseline. This suggests that when the downstream model lacks sufficient capacity to exploit high-dimensional concatenated representations, additional encoders may introduce noise rather than complementary discriminative ability. It is possible that the more advanced gradient boosting models and neural networks may be better able to utilise text encodings from multiple encoders at once to boost predictive ability, but this subject was not rigorously examined as part of this study.

\subsection{Results for Importance of Categorical Features}

The inclusion of one-hot encoded categorical features consistently improved model performance. A statistical analysis combining Chi-squared and Cramer's V,  identified interventionModel ($\chi^2$=219.13, p<0.001), allocation ($\chi^2$=90.88, p<0.001), and oversightHasDmc ($\chi^2$=76.83, p<0.001) as the most statistically significant predictors of dosing error occurrence. From a clinical perspective, this is intuitive: crossover and factorial intervention designs involve more complex dosing schedules and transitions between treatment arms, increasing the procedural risk of dosing errors compared to simpler parallel or single-group designs \citep{england2020patientsafety}. Similarly, the absence of a Data Monitoring Committee may reflect reduced oversight infrastructure, as shown in prior work to be associated with higher rates of protocol deviation and adverse events in clinical trials \citep{stewart2021fda}.

Although the categorical features 'primary purpose' and 'trial phase' did not meet the threshold for statistical significance with regard to impacting the dosage error target feature, domain knowledge supported their inclusion. Their addition produced improvements in several model configurations and did not appear to actively reduce predictive ability, and as such were included in the subsequent experiments. This reflects the broader challenge of feature selection in clinical prediction tasks, where statistical significance does not always align with the predictive ability of machine learning models. 

\subsection{Results for Class Weighting}

A significant methodological challenge in this study was the pronounced class imbalance within the dataset, with approximately 95.4\% of training records labelled as 0 (28,126 of 29,478), indicating that the lower bound of the 95\% Wilson confidence interval for the dosing error rate did not exceed the predefined threshold for those clinical trials. This distribution reflects the real-world rarity of formally reported dosing errors in clinical trial registries, where medication errors are known to be systematically under-reported \citep{delavoipiere2021medication}. Moreover, using ROC-AUC as an evaluation metric can be misleading in highly imbalanced applications, as strong performance may still coincide with poor minority class detection \citep{imani2026rocauc}. To address this, additional evaluation metrics, including precision, recall, and balanced accuracy, were incorporated to ensure that models were meaningfully identifying cases where dosing error likelihood surpassed the defined threshold, rather than defaulting to majority-class predictions.

When using simpler machine learning models such as logistic regression, varying the class weights for the minority class had no meaningful impact on logistic regression performance, with ROC-AUC against the validation set remaining within a narrow range (0.791–0.798) across all weighting strategies (mean: 0.795, standard deviation of 0.003). This suggests that re-weighting did not materially alter ranking-based discriminative performance with respect to ROC-AUC. Alternative approaches, such as upsampling the minority class, may potentially improve sensitivity; however, this was not explored in the present study. In medical contexts, synthetic or repeated sampling strategies must be applied cautiously, as they may introduce over-fitting, distort clinically meaningful distributions, or amplify noise in rare but heterogeneous cases.

In contrast to logistic regression, the neural network demonstrated improved ability to predict dosage error when class weighting was applied, with ROC-AUC increasing from 0.806 (no weighting) to 0.838 using automatically computed class weights based on the proportion of examples of each class within the training data. This indicates that the neural architecture was more responsive to imbalance-aware training, improving its ability to detect minority class instances (dosing error probability surpassed defined threshold) without sacrificing overall discriminative performance (mean: 0.835, standard deviation of 0.002, across experiments where class weighting was applied).

\subsection{Results for Comparison of Classical ML models}
\begin{figure}[ht]
    \centering
    \includegraphics[width=1.0\linewidth]{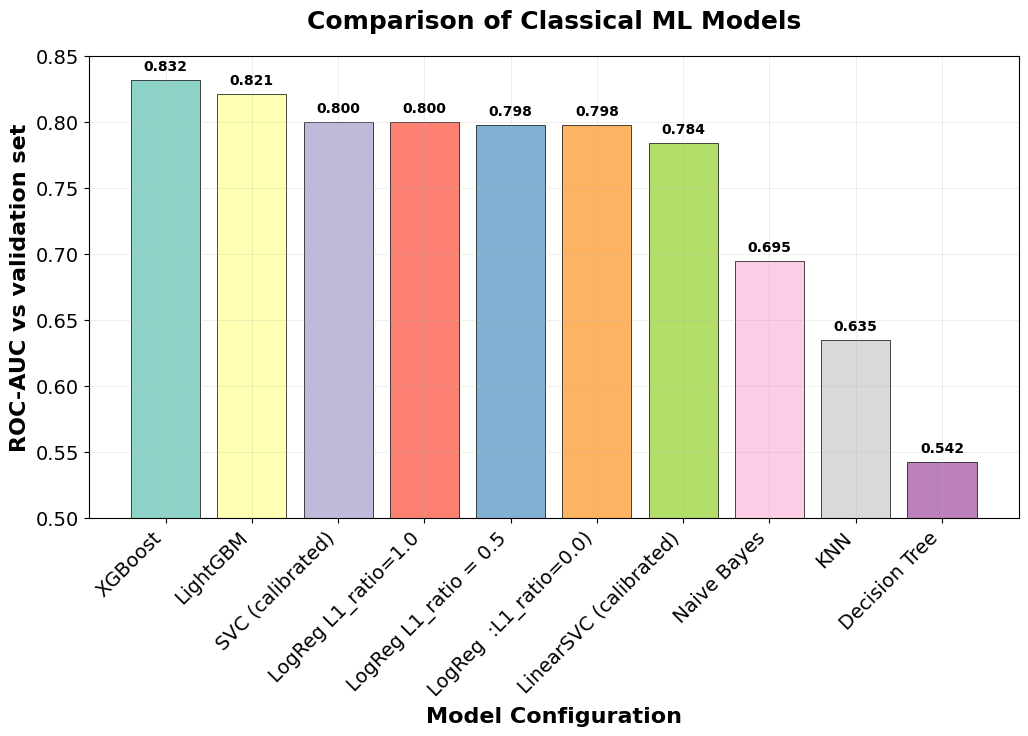}
    \caption{Comparison of Classical ML Models}
    \label{fig 3:comparison classical ML models}
\end{figure}

As shown in Figure~\ref{fig 3:comparison classical ML models}, the strongest discriminative performance on the validation set, as measured by ROC-AUC, was achieved by LightGBM, XGBoost, the calibrated support vector classifier (SVC), and logistic regression. These models consistently outperformed simpler methods such as k-nearest neighbours and decision trees, indicating that either well-regularised linear decision boundaries or gradient-boosted ensemble approaches are particularly well suited to the high-dimensional feature space produced by transformer-based text embeddings combined with categorical variables.

Following light hyperparameter optimisation, LightGBM achieved a validation ROC-AUC of 0.831 using \texttt{num\_leaves = 31}, \texttt{max\_depth = 8}, \texttt{learning\_rate = 0.05}, and \texttt{min\_child\_samples = 50}. Logistic regression achieved a validation ROC-AUC of 0.812 with an L2 penalty and regularisation strength $C = 0.1$ and 'balanced' class weighting. The calibrated SVC achieved a cross-validated ROC-AUC of 0.831 with $C = 1$, $\gamma = 0.001$, and an RBF kernel. These findings suggest that both ensemble-based methods and regularised margin-based classifiers are capable of effectively leveraging the semantic representations generated by domain-specific transformer encoders.

Notably, both tree-based boosting methods and calibrated linear models achieved comparable predictive performance, indicating that substantial gains can be achieved without necessarily increasing architectural complexity. From a practical perspective, logistic regression offers an additional advantage in terms of interpretability. The learned coefficients provide a direct estimate of feature contribution to the predicted probability of a dosing error, enabling transparent inspection of influential variables. When combined with L1 regularisation, or elastic net regularisation with a non-zero L1 ratio, the model promotes sparsity by shrinking less informative coefficients to zero. This facilitates feature selection, reduces model complexity, and enhances interpretability, an important consideration in clinical and regulatory settings where model transparency and justification of decisions are critical.

\subsection{Results for Comparison of Neural Network models}

\begin{figure}[ht]
    \centering
    \includegraphics[width=1.0\linewidth]{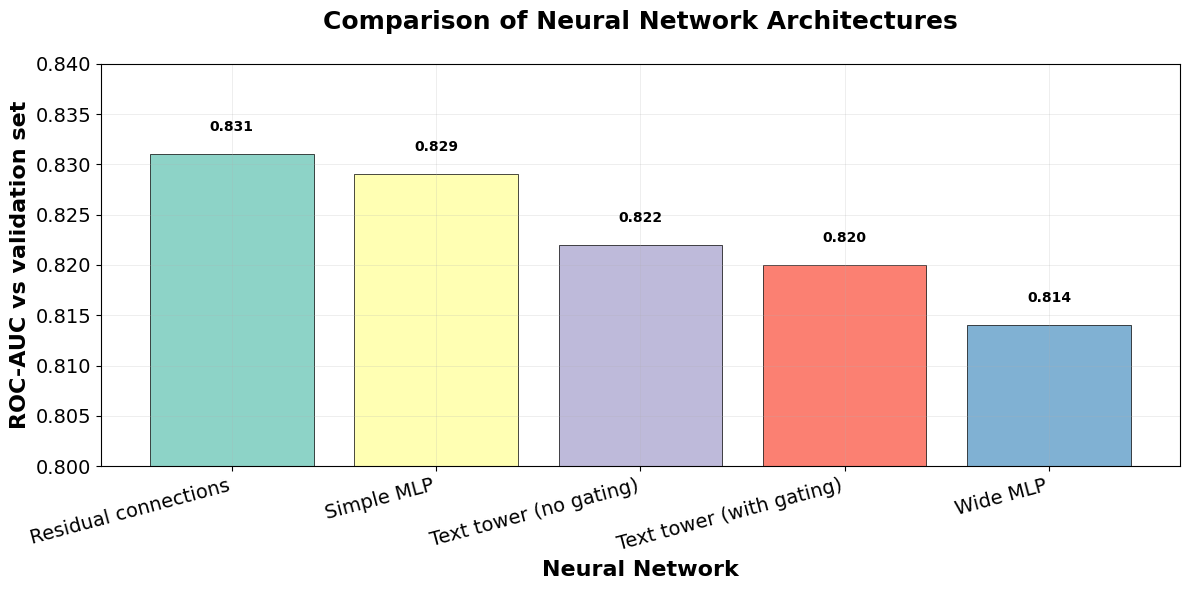}
    \caption{Comparison of Neural Network Models}
    \label{fig 4:comparison neural network models}
\end{figure}

A series of neural network architectures was evaluated to assess the impact of structural design on predictive performance. A simple multi-layer perceptron achieved a validation ROC-AUC of 0.829, providing a strong baseline (see Figure \ref{fig 4:comparison neural network models}). Expanding the architecture into a wider network did not yield improvements, resulting in a lower ROC-AUC  of 0.814, suggesting that increasing parameter count alone did not enhance generalisation. Introducing residual connections led to the strongest performance, with a ROC-AUC  of 0.831, indicating that skip connections may facilitate more effective gradient flow and feature utilisation in the high-dimensional embedding space. Additionally, a text tower architecture, designed to process textual embeddings separately from categorical features before merging, achieved ROC-AUC scores of 0.820 (with gating) and 0.822 (without gating). These results suggest that while architectural modifications influence performance, residual connections provided the most consistent gains, whereas increased width or gated fusion mechanisms within the neural networks did not confer additional benefit.

\subsection{Results after Cross-validation}
Following model selection and hyperparameter optimisation via GridSearch (only applied for classical ML models, specific GridSearch parameters can be seen in Appendix \ref{appendix: Lightgbm Param Optimisation}), a final evaluation was conducted using stratified k-fold cross-validation (k=5) across the combined training and validation datasets. Full fold-level results are reported in Appendix \ref{appendix: k-fold results}.

 The light gradient boosting model, calibrated SVC model, and the text-tower neural network, all demonstrated stronger and more stable discriminative ability across folds. For these models, performance under stratified five-fold cross-validation was consistent with the previous train-validation, indicating stability across sampling splits. The calibrated SVC achieved a mean ROC-AUC of 0.841 ($\pm$ 0.007), while LightGBM achieved 0.836 ($\pm$ 0.010), and the text-tower neural network achieved 0.831 ($\pm$ 0.012). 
 The relatively low standard deviations across folds indicate robustness to sampling variability, suggesting that these models generalise consistently across different subsets of the dataset. Although the calibrated SVC achieved the highest ROC-AUC, its recall at a default decision threshold of 0.5 remained low, reflecting conservative positive class predictions under this operating point.

When comparing performance across the stratified k-fold (k=5) cross-validation data splits against the results when trained with the provided train-validation split, the logistic regression models showed a reduction in mean ROC-AUC under cross-validation. Logistic regression achieved a mean ROC-AUC of 0.732 ($\pm$ 0.006). This decline suggests that earlier performance against the initial training and validation sets may have been bias due to variations in data partitioning. In particular, the reduced performance indicates that this approach may rely more heavily on fold-specific structure within the data, resulting in lower generalisation when evaluated across multiple stratified splits.

All classification metrics were computed using a fixed probability threshold of 0.5 to ensure consistency across models. Given the pronounced class imbalance in the dataset, this threshold may not be optimal for maximising clinically meaningful trade-offs between precision and recall. In practice, threshold tuning based on application-specific priorities (e.g. prioritising sensitivity for early risk detection) could further improve precision, recall, and balanced accuracy without altering the underlying ranking performance as measured by ROC-AUC.

Overall, the cross-validation results provide a more reliable estimate of model generalisation performance and highlight the relative robustness of the calibrated SVC and text-based neural architectures for this task.

\subsection{Summary of Findings}
This study evaluated transformer-based biomedical text embeddings combined with structured categorical features for predicting clinical trial protocols in which the probability of a dosing error exceeded a defined threshold. Model performance was assessed primarily using ROC-AUC, with secondary metrics including F1-macro, precision, recall, and balanced accuracy. Across all experiments, BioBERT consistently outperformed alternative biomedical language models when used as a stand-alone encoder. Furthermore, combining multiple pretrained encoders did not produce additive gains, suggesting that domain alignment was more important than representational stacking. Incorporating structured categorical trial design variables alongside text embeddings improved predictive performance across model classes.

Among the evaluated approaches, residual neural networks, calibrated support vector classifiers, and light gradient achieved the strongest and most stable performance under cross-validation, while gradient boosting models achieved competitive results with lower computational cost. Explicit handling of class imbalance was necessary to obtain meaningful recall of the minority class. These findings highlight the importance of domain-specific pretraining and careful model selection in highly imbalanced clinical prediction tasks.

\section{Conclusion}
This work demonstrates that domain-specific transformer embeddings, particularly BioBERT, provide highly effective representations for modelling unstructured clinical trial documentation. When integrated with structured trial metadata, these embeddings enable reliable discrimination of trials in which the estimated probability of dosing error exceeds a predefined threshold.

The results suggest that encoder-domain alignment plays a greater role in predictive performance than increasing architectural complexity or combining multiple pretrained representations. While neural architectures achieved strong performance, gradient boosting methods offered competitive discriminative ability with greater interpretability and lower computational cost, making them attractive candidates for deployment in regulatory or governance settings where transparency is essential.

Future research may explore threshold optimisation strategies tailored to specific clinical priorities, the application of model explainability techniques to identify high-risk protocol elements, and the extension of this framework to multi-class prediction of dosing error types. Incorporating additional data sources, such as adverse event reports or electronic health records, may further strengthen real-world applicability and support proactive safety monitoring.

\clearpage
\section{Limitations}

Firstly, compute constraints prevented the evaluation of language models with larger amounts of parameters, such as BioGPT \citep{luo2022biogpt} and SciFive \citep{phan2021scifive}. Encodings from these models may have produced stronger semantic representations of clinical trial protocol text. Secondly, the binary target variable conflates dosing errors of varying severity and type into a single label, limiting the granularity of predictions. Future work could explore multi-class formulations that distinguish between error types such as incorrect dose, incorrect frequency, and omission errors, as taxonomised in prior work \citep{england2020patientsafety}. Thirdly, the dataset is derived exclusively from the ClinicalTrials.gov registry, which captures structured protocol information but may not reflect the full complexity of dosing procedures described in supplementary protocol documents or internal trial management systems. As noted in prior work \citep{navar2016openaccess}, registry data is subject to selective reporting and inconsistent documentation practices. 

Finally, while feature importance scores can be extracted from XGBoost and LightGBM, interpreting which specific textual patterns in protocol descriptions drive predictions remains an open challenge that warrants further investigation using explainability techniques such as SHAP values or attention-based visualisation. In addition, the neural network models provide difficulty in explainability of results due to the 'black-box' nature of the neural network's internal learning processes. 

\section{Ethics Statement and Code/Data Availability}

This study utilised publicly available clinical trial registry data and secondary benchmark healthcare datasets via the Hugging Face online platform and was subject to the original data providers’ terms and conditions. Due to licensing and governance restrictions, raw clinical trial and healthcare datasets are not redistributed by the authors. Additionally, the datasets were anonymised in accordance with their respective usage policies. As such, the research did not require additional ethical approval. The study was conducted in accordance with relevant data governance standards and responsible AI research practices.

To facilitate open research and transparency, the code for this study is publicly available via the \href{https://github.com/CaresAI-AU/26_c_ai_ct_deb2026}{GitHub repository}.

\newpage
\section{Bibliographical References}\label{sec:reference}
\bibliographystyle{lrec2026-natbib}
\bibliography{references}

\clearpage
\onecolumn

\appendix
\section{Appendices}

\subsection{Full dataset column description}
\label{appendix: columns}
\begin{longtable}[h]{p{3cm} p{5.5cm} p{6cm}}

\toprule
\textbf{Feature} & \textbf{Feature meaning} & \textbf{Relevance to modelling} \\
\midrule
\endfirsthead
\endhead

phases & Clinical trial phase (e.g. Phase 1–4) & Indicates development stage, regulatory maturity, safety vs efficacy focus, and expected sample size. \\
\hline
enrollmentCount & Number of participants planned or enrolled & Proxy for trial scale, statistical power, and development stage; larger trials typically reflect later phases. \\
\hline
allocation & Method of assigning participants to arms & Core indicator of internal validity and risk of bias; randomisation underpins causal inference. \\
\hline
interventionModel & Structural design of the trial & Determines treatment comparison logic and statistical assumptions. \\
\hline
primaryPurpose & Main intent of the trial & Distinguishes exploratory from confirmatory trials. \\
\hline
masking & Level of blinding & Directly affects bias risk. \\
\hline
healthyVolunteers & Recruitment of healthy participants & Differentiates early-phase from therapeutic trials. \\
\hline
sex & Sex eligibility criteria & Important for generalisability. \\
\hline
oversightHasDmc & Presence of Data Monitoring Committee & Indicates safety oversight rigor. \\
\hline
briefSummary & Plain-language study summary & Useful for NLP-based screening. \\
\hline
detailedDescription & Expanded study description & Rich source of procedural details. \\
\hline
conditions & Diseases studied & Enables disease-specific analysis. \\
\hline
protocolPdfText & Extracted protocol PDF text & Contains detailed operational information. \\
\hline
numArms & Number of study arms & Reflects design complexity. \\
\hline
armDescriptions & Free-text arm descriptions & Encodes dose and schedule information. \\
\hline
interventionNames & Names of interventions & Enables cross-trial drug analysis. \\
\hline
numLocations & Number of study sites & Proxy for geographical spread. \\
\hline
target & Binary dosing error indicator & Supervised learning target. \\
\hline
nctid & Clinical trial identifier & Enables registry linkage. \\

\bottomrule
\addlinespace[0.5em]
\caption{Description of Dataset Features}
\end{longtable}

\subsection{Character and token length of text columns}
\label{appendix:token_length}
\begin{table}[h]
\centering
\label{appendix:char_length_stats}
\begin{tabular}{lrrrrrrrr}
\toprule
\textbf{Field} & \textbf{Count} & \textbf{Mean} & \textbf{Std} & \textbf{Min} & \textbf{25\%} & \textbf{50\%} & \textbf{75\%} & \textbf{Max} \\
\midrule
briefSummary & 29,478 & 491.09 & 459.66 & 16 & 209 & 330 & 592 & 4,900 \\
detailedDescription & 29,478 & 967.18 & 1,505.99 & 0 & 0 & 457 & 1,369 & 30,703 \\
armDescriptions & 29,478 & 347.02 & 496.59 & 0 & 78 & 197 & 425 & 13,350 \\
interventionDescriptions & 29,478 & 282.01 & 332.59 & 17 & 89 & 178 & 347 & 8,745 \\
\bottomrule
\end{tabular}
\caption{Character Length Distribution of Textual Fields}
\end{table}

\begin{table}[h]
\centering
\label{appendix: token_length_stats}
\small
\begin{tabular}{lrrrrrrrrr}
\toprule
\textbf{Field} & \textbf{Mean} & \textbf{Median} & \textbf{Min} & \textbf{Max} & \textbf{Std} & \textbf{25\%} & \textbf{50\%} & \textbf{75\%} & \textbf{\% > 256} \\
\midrule
briefSummary & 117.39 & 81 & 5 & 1,231 & 105.12 & 52 & 81 & 142 & 8.65 \\
detailedDescription & 228.85 & 111 & 2 & 6,917 & 350.95 & 2 & 111 & 327 & 31.81 \\
armDescriptions & 100.12 & 59 & 2 & 4,400 & 138.71 & 27 & 59 & 122 & 7.79 \\
interventionDescriptions & 75.81 & 48 & 6 & 2,297 & 89.17 & 24 & 48 & 94 & 4.03 \\
\bottomrule
\end{tabular}
\caption{Token Length Distribution of Textual Fields Using the BioBERT Tokenizer (Training subset)}
\end{table}

\subsection{Statistical Analysis Categorical Features}
\label{appendix: stat analysis categorical}



\begin{table}[H]
\centering
\small
\begin{tabular}{lccccc}
\toprule
\textbf{Input Feature} & \textbf{$\chi^2$ Statistic} & \textbf{$p$-value} & \textbf{Cramér's V} & \textbf{\makecell{Min Exp. \\Count}} & \textbf{\makecell{$\chi^2$ AND \\Cramér's V}} \\
\midrule
interventionModel   & $2.19 \times 10^{2}$  & $2.28 \times 10^{-45}$  & 0.086 & 5.183   & TRUE  \\
allocation          & $9.09 \times 10^{1}$  & $1.42 \times 10^{-19}$  & 0.056 & 6.926   & TRUE  \\
oversightHasDmc     & $7.68 \times 10^{1}$  & $2.07 \times 10^{-17}$  & 0.051 & 176.854 & TRUE  \\
armGroupTypes       & $1.19 \times 10^{3}$  & $2.53 \times 10^{-26}$  & 0.201 & 0.046   & FALSE \\
phases\_new         & $9.60 \times 10^{2}$  & $5.50 \times 10^{-202}$ & 0.180 & 0.046   & FALSE \\
interventionTypes   & $8.78 \times 10^{2}$  & $1.00 \times 10^{0}$    & 0.173 & 0.046   & FALSE \\
healthyVolunteers   & $1.66 \times 10^{2}$  & $8.73 \times 10^{-37}$  & 0.075 & 1.238   & FALSE \\
primaryPurpose      & $1.25 \times 10^{2}$  & $1.16 \times 10^{-22}$  & 0.065 & 0.138   & FALSE \\
numArms             & 117.849               & $1.17 \times 10^{-13}$  & 0.063 & 0.046   & FALSE \\
masking             & 55.300                & $1.13 \times 10^{-10}$  & 0.043 & 3.211   & FALSE \\
sex                 & 25.387                & $1.28 \times 10^{-5}$   & 0.029 & 0.321   & FALSE \\
\bottomrule
\end{tabular}

\vspace{0.5em}
\caption{Chi-square association analysis and Cramer's V analysis between categorical input features and target variable}
\label{tab:chi_square_analysis}
\end{table}

\subsection{GridSearch - Hyperparameter Search Space}
\label{appendix: Lightgbm Param Optimisation}
\begin{table}[h]
\centering

\begin{tabular}{lll}
\toprule
Model & Hyperparameter & Values \\
\midrule
LightGBM & num\_leaves & \{31, 63\} \\
 & max\_depth & \{6, 8\} \\
 & learning\_rate & \{0.05, 0.1\} \\
 & min\_child\_samples & \{20, 50\} \\
\midrule
Logistic Regression & C & \{0.1, 1, 10\} \\
 & penalty & L2 \\
 & solver & saga \\
\midrule
SVC (RBF) & C & \{0.1, 1, 10\} \\
 & gamma & \{\texttt{scale}, 0.01, 0.001\} \\
 & kernel & RBF \\
\bottomrule
\end{tabular}
\caption{Hyperparameter Search Spaces}
\end{table}

\subsection{Best Models Stratified K-Fold Cross Validation Results}
\label{appendix: k-fold results}
\begin{table}[H]
\centering
\begin{tabular}{lcccccc}
\toprule
Metric & Fold 1 & Fold 2 & Fold 3 & Fold 4 & Fold 5 & Mean $\pm$ Std \\
\midrule
ROC-AUC & 0.728 & 0.722 & 0.740 & 0.737 & 0.733 & 0.732 $\pm$ 0.006 \\
F1 (macro) & 0.570 & 0.557 & 0.567 & 0.568 & 0.563 & 0.565 $\pm$ 0.005 \\
Precision & 0.150 & 0.138 & 0.149 & 0.149 & 0.144 & 0.146 $\pm$ 0.004 \\
Recall & 0.627 & 0.633 & 0.662 & 0.652 & 0.651 & 0.645 $\pm$ 0.013 \\
Balanced Accuracy & 0.728 & 0.722 & 0.740 & 0.737 & 0.733 & 0.732 $\pm$ 0.006 \\
\bottomrule
\end{tabular}
\caption{Stratified k-fold (k=5) Cross-Validation Results — Calibrated SVC}
\end{table}

\begin{table}[H]
\centering
\begin{tabular}{lcccccc}
\toprule
Metric & Fold 1 & Fold 2 & Fold 3 & Fold 4 & Fold 5 & Mean $\pm$ Std \\
\midrule
ROC-AUC & 0.842 & 0.833 & 0.853 & 0.837 & 0.843 & 0.841 $\pm$ 0.007 \\
F1 (macro) & 0.514 & 0.517 & 0.528 & 0.514 & 0.522 & 0.519 $\pm$ 0.005 \\
Precision & 0.429 & 0.526 & 0.500 & 0.375 & 0.429 & 0.452 $\pm$ 0.054 \\
Recall & 0.028 & 0.031 & 0.043 & 0.027 & 0.037 & 0.033 $\pm$ 0.006 \\
Balanced Accuracy & 0.513 & 0.515 & 0.520 & 0.513 & 0.517 & 0.516 $\pm$ 0.003 \\
\bottomrule
\end{tabular}
\caption{Stratified k-fold (k=5) Cross-Validation Results — Logistic Regression}
\end{table}

\begin{table}[H]
\centering

\begin{tabular}{lcccccc}
\toprule
Metric & Fold 1 & Fold 2 & Fold 3 & Fold 4 & Fold 5 & Mean $\pm$ Std \\
\midrule
ROC-AUC & 0.833 & 0.825 & 0.849 & 0.826 & 0.845 & 0.836 $\pm$ 0.010 \\
F1 (macro) & 0.599 & 0.589 & 0.594 & 0.589 & 0.602 & 0.595 $\pm$ 0.005 \\
Precision & 0.181 & 0.169 & 0.175 & 0.169 & 0.183 & 0.175 $\pm$ 0.006 \\
Recall & 0.526 & 0.523 & 0.567 & 0.524 & 0.596 & 0.547 $\pm$ 0.030 \\
Balanced Accuracy & 0.706 & 0.700 & 0.719 & 0.700 & 0.735 & 0.712 $\pm$ 0.013 \\
\bottomrule
\end{tabular}
\caption{Stratified k-fold (k=5) Cross-Validation Results — LightGBM}
\end{table}

\begin{table}[H]
\centering
\begin{tabular}{lcccccc}
\toprule
Metric & Fold 1 & Fold 2 & Fold 3 & Fold 4 & Fold 5 & Mean $\pm$ Std \\
\midrule
ROC-AUC & 0.816 & 0.825 & 0.853 & 0.827 & 0.834 & 0.831 $\pm$ 0.012 \\
F1 (macro) & 0.564 & 0.568 & 0.568 & 0.594 & 0.571 & 0.573 $\pm$ 0.011 \\
Precision & 0.145 & 0.146 & 0.151 & 0.175 & 0.150 & 0.153 $\pm$ 0.011 \\
Recall & 0.639 & 0.563 & 0.701 & 0.549 & 0.612 & 0.613 $\pm$ 0.055 \\
Balanced Accuracy & 0.729 & 0.702 & 0.756 & 0.712 & 0.723 & 0.725 $\pm$ 0.018 \\
\bottomrule
\end{tabular}
\caption{Stratified k-fold (k=5) Cross-Validation Results — Text-Tower Neural Network}
\end{table}

\pagebreak
\subsection{Neural network architectures}
\label{appendix: neural networks}
\subsubsection{Simple Multi-layer Perceptron:}

\begin{lstlisting}[language=Python]
def build_basic_binary_classifier(input_dim):
    model = keras.Sequential([
        layers.Input(shape=(input_dim,)),
        layers.Dense(128, activation="relu"),
        layers.Dropout(0.3),
        layers.Dense(64, activation="relu"),
        layers.Dropout(0.2),
        layers.Dense(1, activation="sigmoid") 
    ])
    return model
\end{lstlisting}

\subsubsection{Wide Multi-layer Perceptron:}
\begin{lstlisting}[language=Python]
def build_binary_classifier_wide(input_dim):
    model = keras.Sequential([
        layers.Input(shape=(input_dim,)),
        
        # Wider first layer to capture feature interactions
        layers.Dense(512, activation="relu", kernel_regularizer=keras.regularizers.l2(0.001)),
        layers.BatchNormalization(),
        layers.Dropout(0.4),
        
        layers.Dense(256, activation="relu", kernel_regularizer=keras.regularizers.l2(0.001)),
        layers.BatchNormalization(),
        layers.Dropout(0.3),
        
        layers.Dense(128, activation="relu", kernel_regularizer=keras.regularizers.l2(0.001)),
        layers.Dropout(0.2),
        
        layers.Dense(64, activation="relu"),
        layers.Dropout(0.2),
        
        layers.Dense(1, activation="sigmoid")
    ])
    return model
\end{lstlisting}

\subsubsection{Residual connections/skip layer:}
\begin{lstlisting}[language=Python]
def build_binary_classifier_skip_layers(input_dim):
    inputs = layers.Input(shape=(input_dim,))
    
    # First block
    x = layers.Dense(512, activation="relu", kernel_regularizer=keras.regularizers.l2(0.001))(inputs)
    x = layers.BatchNormalization()(x)
    x = layers.Dropout(0.4)(x)
    
    # Second block with skip connection
    x2 = layers.Dense(256, activation="relu", kernel_regularizer=keras.regularizers.l2(0.001))(x)
    x2 = layers.BatchNormalization()(x2)
    x2 = layers.Dropout(0.3)(x2)
    
    # Third block
    x3 = layers.Dense(128, activation="relu", kernel_regularizer=keras.regularizers.l2(0.001))(x2)
    x3 = layers.BatchNormalization()(x3)
    x3 = layers.Dropout(0.2)(x3)
    
    # Concatenate different levels (multi-scale features)
    x2_pooled = layers.Dense(64, activation="relu")(x2)
    x3_pooled = layers.Dense(64, activation="relu")(x3)
    combined = layers.Concatenate()([x2_pooled, x3_pooled])
    
    # Final layers
    x = layers.Dense(64, activation="relu")(combined)
    x = layers.Dropout(0.2)(x)
    outputs = layers.Dense(1, activation="sigmoid")(x)
    
    model = keras.Model(inputs=inputs, outputs=outputs)
    return model
\end{lstlisting}

\subsubsection{Text tower model:}
\begin{lstlisting}[language=Python]
def build_single_input_text_cat_fusion_model(
    input_dim: int,
    text_embedding_dim: int = 768,
    num_text_fields: int = 4,
    num_cat_features: int | None = None,  # if None, inferred as input_dim - num_text_fields*text_embedding_dim
    text_proj_dim: int = 192,
    cat_proj_dim: int = 64,
    hidden_dim: int = 128,
    dropout_rate: float = 0.2,
    l2_strength: float = 1e-4,
    use_gating: bool = True,
):
    """
    Single-input model: expects input vector shaped (batch, input_dim)
    Layout: [text1(768) | text2(768) | text3(768) | text4(768) | cat_features(...)]
    """
    l2 = keras.regularizers.l2(l2_strength)

    if num_cat_features is None:
        num_cat_features = input_dim - (num_text_fields * text_embedding_dim)
        if num_cat_features < 0:
            raise ValueError(
                f"input_dim={input_dim} is smaller than {num_text_fields}*{text_embedding_dim}="
                f"{num_text_fields*text_embedding_dim}"
            )

    inputs = layers.Input(shape=(input_dim,), name="flat_input")

    # ---- Slice flat vector into blocks ----
    def slice_block(start_idx, end_idx, name):
        return layers.Lambda(lambda x: x[:, start_idx:end_idx], name=name)(inputs)

    text_blocks = []
    for i in range(num_text_fields):
        start = i * text_embedding_dim
        end = start + text_embedding_dim
        text_blocks.append(slice_block(start, end, name=f"text_{i+1}_slice"))

    cat_start = num_text_fields * text_embedding_dim
    cat_end = cat_start + num_cat_features
    cat_block = slice_block(cat_start, cat_end, name="cat_slice")

    # ---- Per-text-field towers ----
    text_reprs = []
    for i, t in enumerate(text_blocks):
        x = layers.LayerNormalization(name=f"text_{i+1}_ln")(t)
        x = layers.Dense(text_proj_dim, activation="gelu", kernel_regularizer=l2, name=f"text_{i+1}_proj1")(x)
        x = layers.Dropout(dropout_rate, name=f"text_{i+1}_drop1")(x)
        x = layers.Dense(text_proj_dim, activation="gelu", kernel_regularizer=l2, name=f"text_{i+1}_proj2")(x)
        text_reprs.append(x)  # (batch, text_proj_dim)

    # ---- Optional gating over text fields ----
    if use_gating:
        stacked = layers.Concatenate(name="text_concat_for_gate")(text_reprs)  # (batch, num_text_fields*text_proj_dim)
        gate = layers.Dense(64, activation="relu", kernel_regularizer=l2, name="gate_dense")(stacked)
        gate = layers.Dense(num_text_fields, activation="softmax", name="gate_weights")(gate)  # (batch, 4)

        weighted = []
        for i in range(num_text_fields):
            w_i = layers.Lambda(lambda g, idx=i: tf.expand_dims(g[:, idx], axis=-1), name=f"gate_w_{i+1}")(gate)
            weighted.append(layers.Multiply(name=f"text_{i+1}_weighted")([text_reprs[i], w_i]))

        text_fused = layers.Add(name="text_fused")(weighted)  # (batch, text_proj_dim)
    else:
        # simpler: just concatenate all text representations
        text_fused = layers.Concatenate(name="text_fused_concat")(text_reprs)

    # ---- Categorical tower ----
    c = layers.LayerNormalization(name="cat_ln")(cat_block)
    c = layers.Dense(cat_proj_dim, activation="relu", kernel_regularizer=l2, name="cat_proj")(c)
    c = layers.Dropout(dropout_rate, name="cat_drop")(c)

    # ---- Fusion head ----
    x = layers.Concatenate(name="fusion_concat")([text_fused, c])
    x = layers.Dense(hidden_dim, activation="relu", kernel_regularizer=l2, name="fusion_dense1")(x)
    x = layers.Dropout(dropout_rate, name="fusion_drop1")(x)
    x = layers.Dense(hidden_dim // 2, activation="relu", kernel_regularizer=l2, name="fusion_dense2")(x)
    x = layers.Dropout(dropout_rate, name="fusion_drop2")(x)

    outputs = layers.Dense(1, activation="sigmoid", name="output")(x)

    model = keras.Model(inputs=inputs, outputs=outputs, name="single_input_text_cat_fusion")
    return model

\end{lstlisting}
\end{document}